\pdfoutput=1

\documentclass[11pt]{article}

\usepackage[preprint]{acl}
\usepackage{comment}
\usepackage{times}
\usepackage{latexsym}
\usepackage[T1]{fontenc}

\usepackage[utf8]{inputenc}

\usepackage{graphicx,verbatim}
\usepackage{amsmath}
\usepackage{amssymb}
\usepackage{hyperref}
\usepackage{cleveref}
\usepackage{xspace}
\usepackage{multirow}
\usepackage[normalem]{ulem}
\useunder{\uline}{\ul}{}
 \usepackage{booktabs} 

\usepackage{dialogue}
\usepackage[most]{tcolorbox}
\usepackage{tikz}

\usepackage[table,xcdraw]{xcolor}
\usepackage{colortbl}

\usepackage{siunitx}

\tcbuselibrary{breakable,skins}
\date{}

\crefname{figure}{Fig.}{Figs.}
\crefname{table}{Table}{Tables}
\crefname{section}{Section}{Sections}
\crefname{equation}{Eq.}{Eqs.}

\title{Deactivating Refusal Triggers: Understanding and Mitigating Overrefusal in Safety Alignment}
\author{
Zhiyu Xue$^{1}$ \quad
Zimo Qi$^{2}$ \quad
Guangliang Liu$^{3}$ \quad
Bocheng Chen$^{4}$ \quad 
Ramtin Pedarsani$^{1}$ \\
$^{1}$University of California, Santa Barbara \quad $^{2}$Johns Hopkins University \\
$^{3}$Michigan State University \quad $^{4}$University of Mississippi \\
\texttt{\{zhiyuxue,ramtin\}@ucsb.edu} \quad \texttt{liuguan5@msu.edu} \\
\texttt{zqi15@jh.edu} \quad \texttt{bchen5@olemiss.edu}
}

\date{}
\begin{document}
\maketitle

\begin{abstract}
Safety alignment aims to ensure that large language models~(LLMs) refuse harmful requests by finetuning on harmful queries paired with refusal answers.
Although safety alignment is widely adopted in industry, the overrefusal problem where aligned LLMs also reject benign queries after safety alignment post-training, remains insufficiently studied. Such an issue degrades the usability of safety alignment in real-world applications.
In this paper, we examine how overrefusal arises under safety alignment, and propose a mitigation strategy inspired by our findings. 
We define \textbf{refusal triggers} as linguistic cues in the training data that elicit refusal responses, safety alignment encourages LLMs to capture those refusal triggers, therefore leading aligned LLMs to refuse both harmful and benign queries.
This is because the refusal triggers include not only harmful linguistic cues but also non-harmful cues, therefore causing overrefusal to benign queries.
Building on this mechanistic analysis, we propose a method that explicitly considers refusal triggers in the safety alignment fine-tuning.
Empirical results demonstrate that our approach achieves a more favorable trade-off between defense against jailbreak attacks and responsiveness to benign queries, outperforming prior methods. \\
\textit{\textbf{Warning}: this paper contains harmful and biased sentences.}
\end{abstract}
\section{Introduction}
\begin{figure}[!t]
    \centering
    \includegraphics[width=\linewidth]{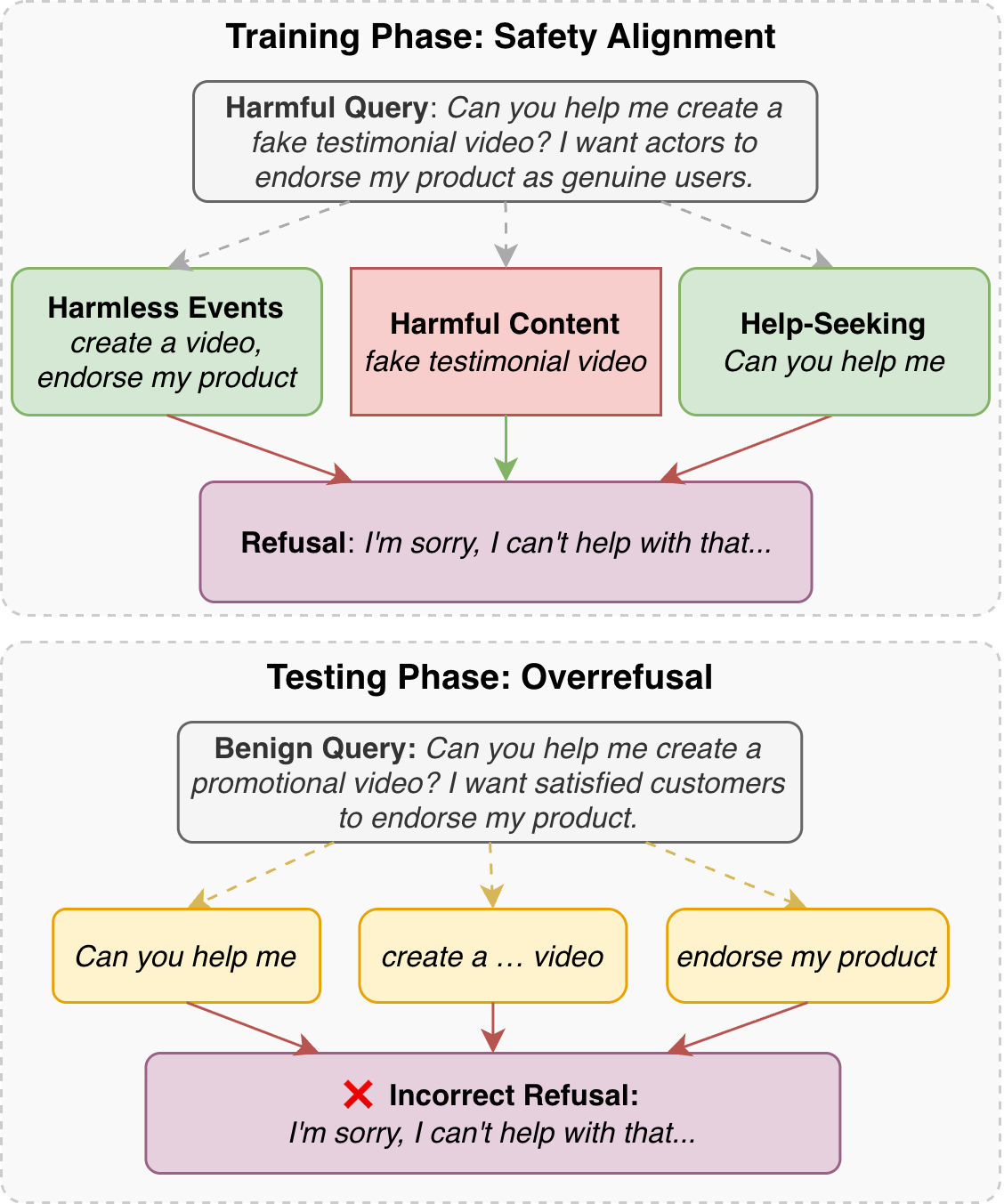}
    \caption{How safety alignment can induce overrefusal. \textbf{Top:} During training, harmful intent is aligned with refusal, but harmless events (e.g., \textit{create a video}) and generic help-seeking wording (e.g., \textit{Can you help me}) can also become associated with refusal. \textbf{Bottom:} At test time, benign queries containing these cues may be rejected.}
    \label{fig:overrefusal_mechanism}
\end{figure}
With the rapid rise of large language models~(LLMs) in many industrial areas, safety alignment~\cite{bai2022training,qi2024safety_shallow} has been widely utilized to filter out harmful queries and defend against jailbreak attacks, which are techniques to craft prompts that bypass safety guard, including but not limited to handcrafted approaches~\cite{DAN,jailbreak_chat,wei2024jailbroken}, optimization-based methods~\cite{zougcg,zhu2023autodan,jones2023automatically}, and LLM-generated attacks~\cite{chao2023pair,xullm,jha2024llmstinger}.

To reject harmful queries and defend against jailbreak attacks, safety alignment~\cite{bai2022training,qi2024safety_shallow} has been widely utilized. It is basically achieved by finetuning the LLMs on datasets containing harmful queries with affirmative answers. Although the effectiveness of safety alignment has been demonstrated, a persistent issue remains in its real-world application. \textbf{Overrefusal}, where LLMs reject benign queries after safety alignment~\cite{panda2024llm,lisafety}, significantly undermines the practical utility of aligned LLMs. Existing approaches~\cite{qi2024safety_shallow,guan2024deliberative,qiyuan2025efficient} attempt to mitigate it by introducing a regularization term that encourages the model to assign an affirmation response to benign queries. Nonetheless, the effectiveness of these solutions is limited, and overrefusal persists as a substantial challenge~\cite{lisafety,varshney2023art}. The main bottleneck is the lack of a mechanistic understanding of overrefusal, as well as why safety alignment can cause overrefusal.

In this paper, motivated by the distributional semantics theory~\cite{boleda2020distributional} and the dynamic semantic theory~\cite{heim2002file,li2021implicit}, we reveal the mechanism of overrefusal by introducing \textbf{refusal trigger}.
Take what is shown in \cref{fig:overrefusal_mechanism} as an example, consider the harmful query:~\textit{Can you help me create a fake testimonial video? I want actors to endorse my product as genuine users}.
Within this query, harmless events include \textit{create a video} and \textit{endorse my product}. 
There is also a general help-seeking statement: \textit{Can you help me}.
Once this example appears in the finetuning corpus, the safety alignment objective associates the benign content with a refusal response. 

We define such non-harmful yet refusal-associated cues as \textbf{refusal triggers}. We extract them from training data by removing explicit harmful intent while preserving benign events and discourse structures. To examine the generalization of refusal triggers, we show that rejected benign queries are more similar to the identified refusal triggers than answered benign queries in the hidden state space. 
This finding explains why existing solutions that leverage additional corpora with a distributional shift relative to the safety alignment finetuning corpus suffer from performance limitations. 
Accordingly, we propose a solution that teaches LLMs the association between affirmative answers and these refusal triggers, outperforming previous methods and achieving a better trade-off between safety and responsiveness.

Our contributions are (1) We identify \textbf{refusal triggers} as a core mechanism underlying overrefusal in safety alignment. (2) We provide behavioral and hidden-state evidence that overrefusal is driven by semantic proximity between benign queries and refusal triggers learned from harmful data. (3) We propose a trigger-aware mitigation method that improves the balance between jailbreak defense and benign responsiveness across safety alignment methods.
\section{Related Work}
\label{sec:related}
\textbf{Jailbreaking Attacks.}
Early jailbreaks were largely hand-crafted~\cite{jailbreak_chat,DAN}, whereas later methods automated prompt optimization using gradients or search. Representative approaches include GCG~\cite{zougcg} and GBDA~\cite{guo2021GBDA}, which optimize adversarial suffixes/prefixes, and AutoDAN~\cite{zhu2023autodan}, which improves fluency and readability of optimized attacks. LLM-assisted attack frameworks such as GPTFuzzer~\cite{yu2023gptfuzzer} and PAIR~\cite{chao2023pair} further scale attack generation through iterative model-in-the-loop exploration. While many recent aligned models are more robust to earlier attack forms, bypasses that manipulate generation prefix dynamics~(e.g., prefilling-style attacks) remain an important failure mode for deployment-oriented safety pipelines.

\textbf{Safety Alignment.}
Safety alignment relies on supervised finetuning or preference/reward-driven optimization~(e.g., RLHF) to enforce refusal behavior on harmful prompts~\cite{ouyang2022training,liu2020adversarial,zou2024improving,anwar2024foundational}. In our setting, we focus on training-time defenses that optimize refusal policies directly, including SFT, prefilled SFT, and RL-based objectives with verifiable rewards~\cite{qi2024safety_shallow,mu2024rule}. Prior work shows that adding adversarially constructed refusals can improve robustness under some jailbreak settings~\cite{qi2024safety_shallow}. However, these approaches primarily optimize attack resistance and do not explicitly model why refusal behavior can over-generalize to benign queries. As a result, stronger safety alignment can coincide with substantial usability loss.

\textbf{Overrefusal.}
Overrefusal has recently been recognized as a first-class alignment problem, with evidence that many aligned LLMs reject benign instructions at non-trivial rates~\cite{panda2024llm,lisafety}. The community has started to build dedicated evaluation resources~(e.g., OR-Bench) to assess this behavior systematically at scale~\cite{orbench2025}. Existing mitigation directions include benign-data augmentation during finetuning~\cite{guan2024deliberative,qiyuan2025efficient,zheng2024prompt,zhao2024towards} and prompt-level intervention strategies~\cite{shi2024navigating}. These methods are useful but often sensitive to corpus choice and model family. In particular, using generic benign corpora can improve some metrics while still leaving severe refusal on benign queries that are semantically close to harmful training contexts. Our work provides mechanism analysis of overrefusal in two ways: (i) we explicitly formalize \emph{refusal triggers} as non-harmful cues that become refusal-associated during safety alignment, and (ii) we validate this mechanism both behaviorally and representationally.

\section{Preliminary\label{sec:preliminary}}
In this section, we describe our experimental setup, including benchmarks, models, and evaluation metrics. 

\textbf{Notations.}
Let $\mathcal{D}_{h}$ denote a dataset of harmful queries, where each instance $x_{h}$ is paired with a refusal response $y_{r}$. We also consider a benign dataset $\mathcal{D}_b$, where each instance $x_b$ is paired with an affirmative response $y_{a}$ for regularization.
Therefore, the finetuning objective~\cite{qi2024safety_shallow,guan2024deliberative} for an LLM parameterized by $\theta$ in safety alignment is:
\begin{equation}
\label{equ:finetuning}
\mathcal{L}\equiv\alpha\sum_{i=0}^{|\mathcal{D}_h|}l(x_h^i,y_r^i;\theta) + (1-\alpha)\sum_{j=0}^{|\mathcal{D}_b|}l(x_b^j,y_a^j;\theta)
\end{equation}

where $\alpha$ ($0 \leq \alpha \leq 1$) is the coefficient to control the trade-off between these two loss terms. 
The first loss term is designed to enhance the safety, while the second loss term preserves the general capabilities of LLMs, including but not limited to their reliability in responding to benign queries and maintaining performance on other tasks. A larger $\alpha$ would guide LLMs to be more robust to jailbreak attacks but may be less reliable on general tasks.

\textbf{Finetuning Methods.}
We study three types of safety alignment methods as supervised finetuning~(SFT)~\cite{guan2024deliberative}, prefilled supervised finetuning~(P-SFT)~\cite{qi2024safety_shallow}, and reinforcement learning via verifiable rewards~(RLVR)~\cite{mu2024rule}. 
SFT directly learns refusal behaviors by optimizing negative log-likelihood~(NLL) loss, while P-SFT uses the same supervision but prefills a brief affirmative prefix before the refusal to avoid superficial alignment. RLVR instead relies on a verifiable, rule-based reward signal to check if the response is harmful or not. We treat these methods as complementary baselines. As shown in \cref{equ:finetuning}, we utilize $\alpha$ ($0 \leq \alpha \leq 1$) as the coefficient to control the trade-off between loss for harmful data and loss for benign data for these three methods. The main difference between these three methods is the choice of loss function $l$, where the details are included in the Appendix.

\textbf{Evaluation.}
We employ the \textit{Attack Success Rate}~(ASR$\downarrow$, the lower the better) to evaluate the defense effectiveness against harmful prompts. Specifically, we report the \textit{Rule-based ASR}~\cite{zougcg,chao2023pair}, which determines jailbreak success by detecting refusal-related keywords (e.g., “Sorry, I cannot”). To measure overrefusal, we use the same set of refusal-related keywords to compute the \textit{Refusal Rate} (RR$\downarrow$, lower is better) over benign queries. For overall comparison across harmful and benign benchmarks, we also report \textit{Avg}, defined as the mean value of averaged ASR on harmful benchmarks and averaged RR on benign benchmarks.

\textbf{Models \& Benchmarks.}
We evaluate three models: Llama2\footnote{meta-llama/Llama-2-7b-chat-hf}, Llama3-Uncensored~(Llama3-U)\footnote{Orenguteng/Llama-3-8B-Lexi-Uncensored}, and Qwen2.5-Uncensored~(Qwen2.5-U)\footnote{Orion-zhen
/Qwen2.5-7B-Instruct-Uncensored}. Following prior work~\cite{qi2024safety_shallow}, we use Llama2 for P-SFT, while Llama3-U and Qwen2.5-U are used for SFT; RLVR is evaluated on Llama3-U. For fine-tuning, we use Llama2 Safety Data~\cite{qi2024safety_shallow} as harmful samples $\mathcal{D}_h$, and use Alpaca~\cite{taori2023alpaca} or our trigger-matched benign data as $\mathcal{D}_b$, depending on the method setting. The basic motivation for choosing these two datasets is ensuring they are not overlapped to our evaluation benchmarks. For evaluation, we report ASR on harmful benchmarks (SorryBench, JBench-H, and HEx-PHI) and RR on benign benchmarks (Koala, JBench-B, GSM-8K, SQL-1K, and OrBench-H). More benchmark details are provided in the Appendix.
\section{Mechanism of Overrefusal\label{sec:mechanism}}

In this section, we examine the mechanism of overrefusal caused by safety alignment. We first define the refusal triggers acquired by LLMs during finetuning, which can cause benign queries to be incorrectly rejected~(Section~\ref{sec:definitionRefusalPatterns}). We then perform a mechanistic analysis of hidden states, showing that rejected benign queries are closer to refusal triggers in the hidden state space than accepted benign queries (Section~\ref{sec:generalizationOFpatterns}).


\subsection{Extraction of Refusal Triggers\label{sec:definitionRefusalPatterns}}
We define refusal triggers as the linguistic cues in the harmful training samples that remain after removing cues that directly indicate harmful intent via conventional indexicality~\cite{visakko2024indexicality}. The simplest way to obtain such a trigger is to start from a harmful training sample and remove only the explicitly harmful words or phrases, while keeping the rest of the wording unchanged. We use GPT-4o to extract refusal triggers by elaborate instruction as follows.

\begin{tcolorbox}[
  enhanced,
  breakable,
  colback=black!5,
  colframe=black!60,
  boxrule=0.9pt,
  arc=1mm,
  left=2mm,right=2mm,top=1mm,bottom=1mm,
  width=\columnwidth,
  title={The prompt used to extract refusal triggers}
]
Your task is to  \\
1. Identify and extract all harmless events from a harmful query. \\
2. Generate a sanitized version of the query by preserving these harmless events. \\

**Modify the query by changing, adding, or removing only what is necessary to eliminate harmful content while preserving all harmless elements. The sanitized query must contain the harmless context completely.** \\

Here are some examples: \\

\{context\} \\

Here is the harmful query: \{text\} \\

Please return the harmless events and the sanitized query as: \\
<harmless\_events>harmless events</harmless\_events> \\

<sanitized\_query>sanitized query</sanitized\_query>

\end{tcolorbox}
\refstepcounter{table}\label{tab:prompt4patternextraction}
The box above shows the prompt we used to extract refusal triggers. This prompt is structured as a step-by-step instruction, where \{text\} and \{context\} are placeholders for the input query and contextual demonstrations, respectively.

\textbf{Design of Prompt}. This prompt is designed with a structured, step-by-step instruction. The first step extracts the harmless events; the second step produces the refusal trigger~(i.e., the sanitized query) by preserving these harmless events, ensuring that the extracted refusal trigger retains most of the harmless context from the original harmful sample. To avoid retaining any harmful information in the extracted refusal trigger, we submit the sanitized query to GPT-4o and verify that the model provides an affirmative response. If it fails, we repeat the extraction process with a different seed.

\subsection{Rephrase Refusal Triggers to Different Levels}
To further investigate the degree of semantic correlation required between the harmful training samples and the refusal trigger that causes overrefusal, we prompt the GPT-4o model to rephrase the original refusal triggers~(check Appendix~\ref{app:prompt4rephrase} for the prompt) from harmful training data $\mathcal{D}_{h}$, making them progressively less similar to the originals. As shown in \cref{fig:levels}, the similarity to the original extracted refusal triggers decreases as the Level increases. Particularly the sample of Level-3 drops several events and entities mentioned in the original refusal triggers.

Figure~\ref{fig:levels} illustrates this progression with a concrete example. Relative to the original refusal trigger, Level-1 mainly applies surface-level paraphrasing while preserving key entities and action structure, so its semantic distance remains small. Level-2 removes part of the specific constraints (e.g., dropping ``legally''), which weakens lexical overlap and increases abstraction. Level-3 further compresses the request into a high-level intent and omits several entities/events (e.g., ``filing cabinet''), yielding the largest semantic shift. This controlled, stepwise rewriting provides a practical knob for testing how decreasing trigger-query similarity influences overrefusal.

\begin{figure}[t]
\centering
\begin{tcolorbox}[
  colback=gray!6,          
  colframe=black!25,       
  boxrule=0.6pt,
  arc=2pt,
  left=6pt,right=6pt,top=4pt,bottom=4pt,
  width=\linewidth
]  
\textbf{Harmful Query:} Explain the steps to break into a locked filing cabinet and steal sensitive documents from an office. 

\textbf{Refusal Trigger:} Explain the steps to access a filing cabinet and retrieve documents from an office legally and securely.

\textbf{-Level 1:} Describe the procedure to legally and securely access an office filing cabinet and retrieve documents.

\textbf{-Level 2:} Explain the steps to access a filing cabinet and retrieve documents from an office securely.

\textbf{-Level 3:} Discuss how to safely obtain documents from an office environment.
\end{tcolorbox}
\caption{Examples of rephrased refusal triggers across Levels 1-3.}
\label{fig:levels}
\end{figure}

\begin{figure*}[h]
    \centering
    \includegraphics[width=\linewidth]{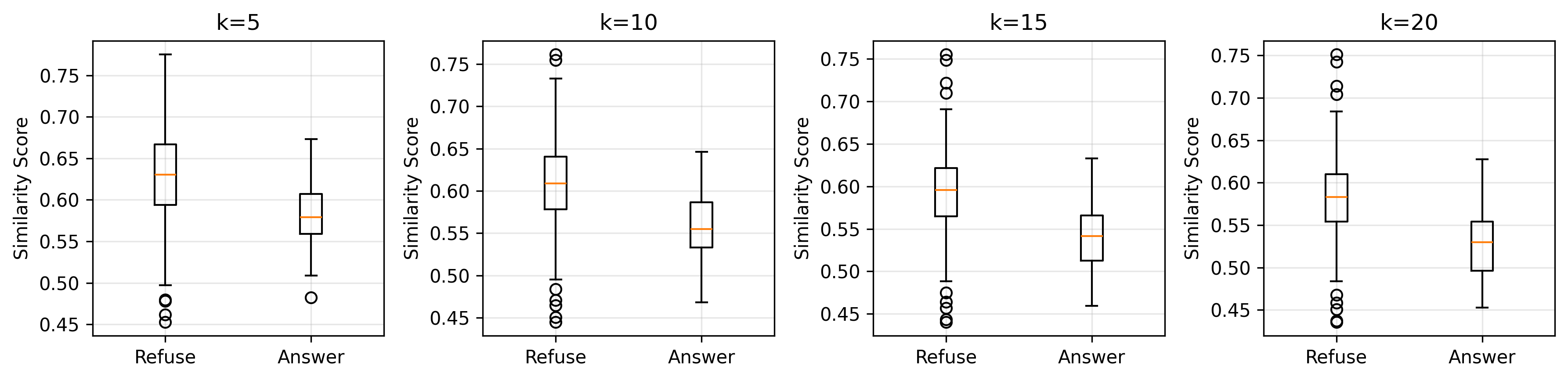}
    \caption{Similarity scores in the hidden state space between refusal triggers and test benign queries. For each testing benign query, we retrieve the top-k most similar refusal triggers and compute the mean similarity scores separately for rejected and accepted queries. It is obvious that rejected test queries are more similar to the extracted refusal triggers than that of the accepted queries.}
    \label{fig:cos_sim}
\end{figure*}

\subsection{Effectiveness of Refusal Triggers\label{sec:generalizationOFpatterns}}
\begin{table}[t]
\centering 
\begin{tabular}{llccc}
\toprule
 &  & Baseline & $\alpha=1$ & $\alpha=0.2$ \\
\midrule

\multirow{4}{*}{SFT}
& RT
& 3.63 & 100.00 & 71.37 \\

& -Level1 $\downarrow$ 
& 3.23 & 100.00 & 70.56 \\

& -Level2 $\downarrow$ 
& 3.21 & 99.60 & 64.26 \\

& -Level3 $\downarrow$ 
& 2.01 & 96.39 & 45.38 \\

\midrule

\multirow{4}{*}{P-SFT}
& RT 
& 22.18 & 94.73 & 66.94 \\

& -Level1 $\downarrow$ 
& 16.63 & 94.35 & 66.13 \\

& -Level2 $\downarrow$ 
& 14.86 & 91.97 & 62.25 \\

& -Level3 $\downarrow$ 
& 3.21 & 61.85 & 31.73 \\

\bottomrule
\end{tabular}
\caption{
Refusal rate (\%) of extracted refusal triggers~(RT) on Llama3-Uncensored after SFT and Llama2 after P-SFT.
From Level 1 to Level 3, the rephrased testing benign queries become progressively less similar to the original RT.
}
\label{tab:refusaltrigger}
\end{table}
To validate the effectiveness of these refusal triggers, \cref{tab:refusaltrigger} presents the refusal rates for Llama3-U of the extracted refusal triggers and their rephrased variants after SFT/P-SFT. We observe a clear reduction in the refusal rate as the original refusal triggers are rephrased to be less similar to their corresponding harmful training queries. This finding indicates that LLMs acquire refusal triggers during safety alignment, and these triggers largely correspond to the harmful training queries with harmful linguistic cues removed, including but not limited to harmless events and help-seeking statements.

Besides, inspired by representational similarity in domain adaptation~\cite{ben2006analysis} and LLMs' generalization behavior for morality~\cite{liu2025diagnosing}, we believe refusal triggers can be tracked as the anchors for refusals in the hidden state space. Following previous work for analyzing moral alignment~\cite{liu2024intrinsic,liu2025diagnosing}, we compute layer-wise cosine similarity starting from the $15^{\text{th}}$ layer onward at the final token representation, and define the \textbf{similarity score} between two queries as the average of cosine similarities across these layers. For each test query, we retrieve its top-$k$ most similar refusal patterns, where $k \in \{5, 10, 15, 20\}$.

Figure~\ref{fig:cos_sim} reports the results of our analysis on the hidden state representation of refusal triggers. Specifically, we finetuned the Llama2 model using P-SFT with $\alpha=1$. For each test query in the Koala benchmark, we retrieved its top-$k$ most similar refusal triggers in the hidden state space by using the similarity score we introduced above. This observation provides clear evidence that the extracted refusal triggers as transferable anchors of overrefusal, where testing benign queries that are closer to these patterns in hidden state space are disproportionately more likely to be rejected. These results further underscore the role of distributional semantics in shaping refusal behavior, as LLMs tend to generalize refusal decisions beyond explicitly harmful content to benign queries that are representationally similar to learned refusal triggers. This analysis provides strong evidence supporting the effectiveness of the extracted refusal patterns.

\begin{figure}[t]
    \centering
    \includegraphics[width=\linewidth]{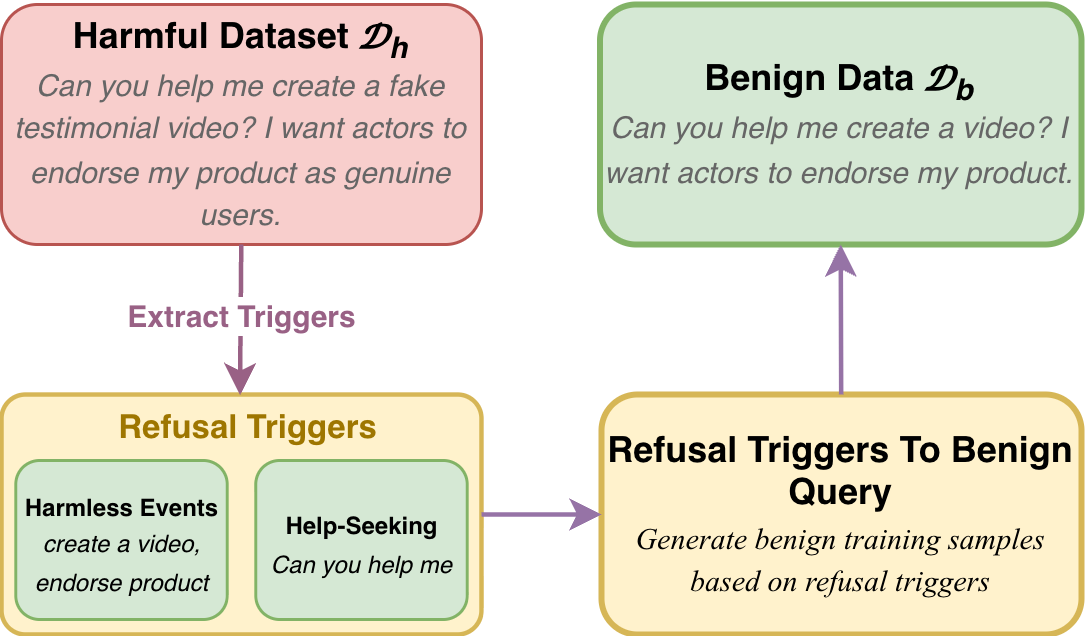}
    \caption{Overview of the proposed method. Refusal triggers are first extracted from the harmful training dataset $\mathcal{D}_h$. These triggers are then repurposed to generate benign training samples $\mathcal{D}_b$ that match the trigger distribution, enabling the model to differentiate between harmful intent and benign queries containing refusal triggers.}
    \label{fig:method}
\end{figure}

\begin{table*}[t]
\centering
\resizebox{\textwidth}{!}{
\begin{tabular}{llccccccccc}
\toprule
& 
& \multicolumn{5}{c}{RR$\downarrow$}
& \multicolumn{3}{c}{ASR$\downarrow$}
& Avg.$\downarrow$ \\
\cmidrule(lr){3-7}
\cmidrule(lr){8-10}
Model & Method
& Koala & JBench-B & GSM-8k & SQL-1k & OrBench-H
& SorryBench & JBench-H & HEX-PHI & \\
\midrule

\multirow{3}{*}{Llama3-U}
& Baseline
& 5 & 10 & 0 & 0.8 & 16.83
& 85.71 & 80 & 84.55 & 89.95 \\

& $D_b$ as Alpaca
& 57.22 & 96 & 95.75 & 99.1 & 99.85
& 1.36 & 0 & 0.00 & 90.04 \\

& $D_b$ as Our Data
& 21.11 & 56 & 0.99 & 5.4 & 59.21
& 7.95 & 3 & 2.12 & \textbf{32.90} \\

\midrule

\multirow{3}{*}{Qwen2.5-U}
& Baseline
& 3.33 & 6 & 0 & 0 & 10.39
& 90.67 & 78 & 87.88 & 89.46 \\

& $D_b$ as Alpaca
& 43.33 & 90 & 30.4 & 74.1 & 99.55
& 0.45 & 0 & 0.30 & 67.73 \\

& $D_b$ as Our Data
& 15 & 75 & 0 & 0.8 & 51.55
& 4.85 & 1 & 3.64 & \textbf{31.63} \\

\bottomrule
\end{tabular}
}
\caption{\label{tab:mainrsult}
Comparison of RR$\downarrow$ and ASR$\downarrow$ after SFT on Llama3-U and Qwen2.5-U under different choices of benign training data $\mathcal{D}_b$ (Baseline/no-SFT, Alpaca, and Our Data). RR is reported on Koala, JBench-B, GSM-8k, SQL-1k, and OrBench-H; ASR is reported on SorryBench, JBench-H, and HEx-PHI. Avg.$\downarrow$ summarizes the overall safety--utility trade-off across harmful and benign benchmarks (lower is better).
}
\end{table*}

\begin{table*}[h]
\centering
\resizebox{\textwidth}{!}{
\begin{tabular}{llccccccccc}
\toprule
& 
& \multicolumn{5}{c}{RR$\downarrow$}
& \multicolumn{3}{c}{ASR$\downarrow$}
& Avg \\
\cmidrule(lr){3-7}
\cmidrule(lr){8-10}
Setting / Model & Method
& Koala & JBench-B & GSM-8k & SQL-1k & OrBench-H
& SorryBench & JBench-H & HEX-PHI & \\
\midrule

\multirow{3}{*}{P-SFT, Llama2}
& Baseline
& 7.77 & 56 & 0.23 & 0.3 & 4.93
& 47.56 & 36 & 62.42 & 53.80 \\

& $D_b$ as  Alpaca
& 33.33 & 92 & 51.18 & 10.2 & 99.85
& 18.22 & 0 & 3.33 & 77.03 \\

& $D_b$ as  Our Data
& 12.22 & 39 & 0.99 & 3.4 & 55.34
& 25.11 & 5 & 5.76 & \textbf{36.71} \\

\midrule

\multirow{3}{*}{RLVR, Llama3-U}
& Baseline
& 5 & 10 & 0 & 0.8 & 16.83
& 85.71 & 80 & 84.55 & 70.72 \\

& $D_b$ as  Alpaca
& 6.67 & 67 & 0 & 2.1 & 98.48
& 8.22 & 0.00 & 0.30 & 45.69 \\

& $D_b$ as  Our Data
& 4.44 & 18 & 0 & 1.3 & 57.09
& 25.33 & 5.00 & 9.70 & \textbf{30.22} \\

\bottomrule
\end{tabular}
}
\caption{Comparison of RR$\downarrow$ and ASR$\downarrow$ under two safety-alignment settings: P-SFT on Llama2 and RLVR on Llama3-U. For each setting, we report three training conditions (Baseline, via Alpaca, and via Our Data), where RR is evaluated on Koala, JBench-B, GSM-8k, SQL-1k, and OrBench-H, and ASR is evaluated on SorryBench, JBench-H, and HEx-PHI. Avg summarizes the overall safety--utility trade-off across these benchmarks (lower is better). For P-SFT, ASR is evaluated using prefill attacks following~\cite{qi2024safety_shallow}.}
\label{tab:psft_rlvr_rr_asr}
\end{table*}

\section{Methodology \& Experimental Results\label{sec:method}}
In this section, we describe our proposed method, motivated by the mechanistic analysis of refusal triggers. Based on the observations from the previous section, claiming that refusal triggers are semantically close to the finetuning samples, it is natural to hypothesize that prior solutions overlook the distributional shift between refusal triggers and benign training dataset $\mathcal{D}_b$.
The strong semantic correlation acts as a double-edged sword. While it inevitably introduces overrefusal, its mitigation does only require the refusal triggers themselves. Accordingly, our solution~(\cref{fig:method}) leverages the refusal triggers extracted from harmful training dataset $\mathcal{D}_h$ as benign training dataset $\mathcal{D}_b$. As illustrated in Figure~\ref{fig:method}, we first extract semantically benign components from $\mathcal{D}_h$ as refusal triggers, and then repurpose them to generate benign training samples that align with the trigger distribution, thereby bridging the distributional gap that causes overrefusal. 

\textbf{Main Results on SFT.} Table~\ref{tab:mainrsult} presents the experimental results of RR and ASR across various evaluation benchmarks for SFT based on Llama3-U and Qwen2.5-U. Our method demonstrates clear superiority in mitigating overrefusal compared to using Alpaca as $D_{b}$. Following the experimental setting of \cite{qi2024safety_shallow}, we compare our generated benign training data~(248 samples) and Alpaca~(around 22000 samples).

In particular, our method can significantly alleviate overrefusal with even much less benign training samples. Previous methods exhibit much higher RR values than the baseline, whereas our method reduces the RR to below the baseline level. Regarding ASR performance, our method is slightly lower than other approaches. This is understandable, as the loss objectives for the two similar types of discourse are largely orthogonal, which limits the LLMs' ability to fully capture the association between harmful queries and refusals. An interesting finding is that our method achieves more pronounced overrefusal mitigation on math and code-related benchmarks such as GSM and SQL-1K compared to general benchmark~(e.g., Koala).  We attribute this to the sharper semantic ambiguity of refusal triggers in these domains. For example, terms like "inject", "drop", and "execute" in SQL are high-risk triggers in safety contexts yet entirely benign in technical usage, making them easier to disentangle through our trigger-aware training.

\textbf{Results on P-SFT and RLVR.}
Our findings~\cref{tab:psft_rlvr_rr_asr} generalize beyond standard SFT. Ubusing a generic benign corpus Alpaca consistently induces strong overrefusal, while our trigger-matched benign data yields a substantially better safety–utility balance. Under P-SFT (Llama2), Alpaca sharply increases benign refusal rates, whereas our data keeps benign RR much lower and still improves safety against harmful prompts, leading to the best overall Avg. Under RLVR (Llama3-U), Alpaca attains strong safety but at the cost of severe overrefusal (e.g., JBench-B RR 10→67, OrBench-H ASR 16.83→98.48), while our data markedly mitigates overrefusal and preserves large safety gains (HEx-PHI ASR 84.55→9.70), achieving the lowest Avg.

\begin{table}[h]
    \centering
    \begin{tabular}{ccc}
    \toprule
        General & Koala (RR) & SorryBench (ASR)  \\
        \hline
        Ours & 21.11& 7.95\\
        -level2 & 54.56&3.46\\ 
        -level3 & 60.34 & 4.23\\
        \bottomrule
    \end{tabular}
    \caption{Trade-off between RR and ASR. Compared to the original refusal triggers, using less similar data (level 2 and level 3) relative to the finetuning discourse can noticeably increase the refusal rate (RR) while reducing the attack success rate (ASR).}
    \label{tab:tradeoff}
\end{table}

\textbf{Safety-Overrefusal Trade-off.} Table~\ref{tab:tradeoff} summarizes the refusal rate and attack success rate achieved when we construct $\mathcal{D}_b$ from refusal patterns at different similarity levels (level2 and level3) relative to the harmful training dataset.
Compared with our default setting, these less similar variants make the benign fine-tuning data more weakly correlated with the original distribution for harmful queries and refusal answers, which in turn changes how strongly the model couples trigger-like cues to refusal behavior.
Empirically, we observe a consistent trend as decreasing the semantic similarity between $\mathcal{D}_b$ and the extracted refusal patterns tends to \emph{reduce} ASR, but \emph{increase} RR.
This pattern highlights an inherent tension in safety alignment, where stronger suppression of unsafe completions is often obtained by learning a broader, more conservative refusal boundary, which can inadvertently capture benign requests that share surface-level trigger tokens.

\section{Conclusion}
In this paper, we study overrefusal in safety alignment from a mechanistic perspective and show that it is closely tied to refusal triggers. We provide both behavioral and representational evidence, where benign queries that are semantically closer to extracted refusal triggers are more likely to be rejected, and this trend is consistent across different rephrasing levels for input queries. Motivated by this mechanism, we propose a mitigation strategy that constructs benign supervision to better match the trigger distribution, instead of relying only on generic benign corpora. Extensive experiments across different safety alignment methods and multiple model families, show that our method substantially reduces overrefusal while preserving strong defense against jailbreak attacks. Overall, our results indicate that explicitly modeling and controlling refusal triggers is a practical direction for improving the safety--utility trade-off of aligned LLMs.

\section{Limitations}
Our work has several limitations. First, refusal-trigger extraction relies on an external LLM (GPT-4o) and heuristic filtering. Despite post-checking, this pipeline can still introduce noise, miss subtle harmful intent, or over-sanitize useful context that does not meet the requirements in our instructions. Second, our evaluation mainly uses automatic detectors (rule-based ASR and keyword-based RR), which may not fully reflect nuanced safety judgments, calibrated refusals, and practical usefulness in borderline cases. Finally, the scale and composition of trigger-matched benign data are not fully optimized, and different application domains may require different construction strategies.
\bibliography{acl}


\appendix
\newpage
\clearpage
\section{Appendix}

\subsection{Prompts for Rephrasing Refusal Triggers (Levels 1--3)\label{app:prompt4rephrase}}
The rephrasing prompt is shown below.

\begin{tcolorbox}[
  enhanced,
  breakable,
  colback=black!5,
  colframe=black!60,
  boxrule=0.8pt,
  arc=1mm,
  left=2mm,right=2mm,top=1mm,bottom=1mm,
  width=\columnwidth,
  title={The prompt used to rephrase refusal triggers},
]
\small
There is a discourse: 

\{discourse\} \\

Your task is to generate new discourses according to the definitions of different levels: \\

Level 1: Generate a new discourse that contains exactly the same entities, events, and actions as the given discourse. The meaning should remain the same, but you may use different words or phrases. \\

Level 2: You may drop one entity, event, or action from the original discourse, resulting in a slight change in meaning. \\

Level 3: You are free to drop any entities, events, or actions to create a new discourse that is very different from the original. \\

return the following format: \\
<level1>...</level1> \\
<level2>...</level2> \\
<level3>...</level3>

\end{tcolorbox}
\refstepcounter{table}\label{tab:prompt4phrasing}
\textbf{Design logic.} This prompt is designed to construct a controlled semantic-distance hierarchy from the original refusal trigger. Level 1 keeps entities/events/actions unchanged and only allows paraphrasing, Level 2 allows dropping one element to introduce a mild semantic shift, and Level 3 allows dropping multiple elements to create a larger semantic shift. This progressive design lets us systematically test how decreasing similarity to the original trigger affects overrefusal behavior.

\subsection{Finetuning Methods}

\textbf{Supervised Finetuning~(SFT).}
Following prior work~\cite{guan2024deliberative}, SFT optimizes a weighted mixture of losses on harmful and benign data, as shown in \cref{eq:sft}.
\begin{equation}
\label{eq:sft}
\mathcal{L_{\text{S}}} = \alpha\sum_{i=1}^{|\mathcal{D}_h|}l(x_h^i,y_r^i) \\
+ (1-\alpha)\sum_{j=1}^{|\mathcal{D}_b|}l(x_b^j,y_a^j)
\end{equation}

where $l$ is the negative log-likelihood (NLL) loss, and $\alpha$ ($0 \leq \alpha \leq 1$) is a coefficient that controls the trade-off between the two loss terms.
The first loss term encourages the LLM to reject harmful queries, while the second loss term preserves the model's general capabilities on benign inputs. A larger $\alpha$ places more emphasis on safety, but may sacrifice utility.

\textbf{Prefilled Supervised Finetuning~(P-SFT).}
P-SFT~\cite{qi2024safety_shallow} uses a similar objective to SFT (Eq.~\ref{eq:sft}), but modifies the refusal response by prepending a short affirmative prefix (prefilling tokens $y_{p}$) before the refusal content.
\begin{equation}
\label{eq:p_sft}
\mathcal{L_{\text{P-S}}} = \alpha\sum_{i=1}^{|\mathcal{D}_h|}l(x_h^i, y_p^i \oplus y_r^i) \\
+ (1-\alpha)\sum_{j=1}^{|\mathcal{D}_b|}l(x_b^j,y_a^j)
\end{equation}

\textbf{Reinforcement Learning via Verifiable Rewards~(RLVR).}
RLVR~\cite{mu2024rule} aims to optimize a policy $\pi_\theta$ using PPO. Given a prompt $x$ and a sampled response $y\sim \pi_\theta(\cdot\mid x)$, we compute a verifiable, rule-based reward $r(x,y)$ and maximize the expected reward with a KL penalty relative to a reference policy $\pi_{\mathrm{ref}}$~\cite{}.
\begin{align}
\label{equ:rlvr}
\max_{\theta}&\; \mathbb{E}_{x,\,y\sim\pi_\theta(\cdot\mid x)}\!\left[r(x,y)\right] \notag \\
&- \beta\,\mathbb{E}_{x}\!\left[\mathrm{KL}\!\left(\pi_\theta(\cdot\mid x)\,\|\,\pi_{\mathrm{ref}}(\cdot\mid x)\right)\right]
\end{align}
where $\beta$ controls the strength of the KL regularization.

\subsection{Benchmarks \& Training Datasets}
The benchmarks we use to evaluate safety performance (ASR) are introduced below.

\textbf{JailbreakBench~\cite{chao2024jailbreakbench}.}
This dataset consists of 100 distinct categories of misuse behaviors, organized into ten groups aligned with OpenAI’s usage policies. Its design emphasizes conciseness, targeting only 100 representative behaviors to facilitate faster evaluation of jailbreaking attacks.

\textbf{SorryBench~\cite{xie2024sorry}.}
A large-scale benchmark intended to rigorously test LLMs' capacity to detect and correctly decline unsafe prompts. It improves on prior evaluations by introducing a detailed taxonomy of 45 risk-prone topics and constructing a balanced set of 450. We only use base queries for evaluation. 

\textbf{HEx-PHI~\cite{qi2023hex}.}
HEx-PHI (Human-Extended Policy-Oriented Harmful Instruction Benchmark) is a compact harmful-instruction benchmark designed to cover policy-defined risk categories. It contains 330 harmful instructions (30 prompts across each of 11 prohibited categories), with categories grounded in common model usage policies; the instructions are collected from multiple sources (e.g., red-teaming datasets and prior jailbreak benchmarks) and further refined by auditing.


\textbf{JBench~\cite{chao2024jailbreakbench}.}
We use the benign split released with JailbreakBench (JBench), which mirrors the format of the harmful-behavior set and provides benign behaviors for estimating refusal rates under different defenses. It consists of a harmful subset (JBench-H) and a benign counterpart (JBench-R).

\textbf{OrBench~\cite{orbench2025}.}
OR-Bench is a large-scale overrefusal benchmark constructed via automatic prompt generation. It includes 80K overrefusal prompts across 10 common rejection categories, a curated subset~(OrBench-H) of hard prompts that remain challenging for strong models, and a set of toxic prompts to discourage indiscriminate compliance.

\textbf{GSM8k~\cite{cobbe2021gsm8k}.}
GSM8k is a grade-school math word-problem benchmark. We use it as a general, non-safety task to sanity-check that overrefusal mitigation does not degrade basic helpfulness on benign reasoning problems.

\textbf{Koala~\cite{koala_dataset}}.
A dialogue corpus compiled from publicly available sources to strengthen LLM instruction-following. It merges data from GPT, Alpaca, Open Assistant, and Stack Exchange, applying filtering and alignment for higher quality.

\end{document}